\documentclass{article}
\usepackage{ijcai22}

\usepackage{times}
\usepackage{soul}
\usepackage{url}
\usepackage[hidelinks]{hyperref}
\usepackage[utf8]{inputenc}
\usepackage[small]{caption}
\usepackage{graphicx}
\usepackage{amsmath}
\usepackage{amsthm}
\usepackage{booktabs}
\usepackage{algorithm}
\usepackage{algorithmic}
\usepackage{amsmath}
\DeclareMathOperator*{\argmax}{argmax}
\usepackage{amsfonts}
\usepackage{bm}
\usepackage{multirow}
\usepackage{multicol}
\usepackage[detect-all]{siunitx}
\usepackage{etoolbox}
\robustify\bfseries
\usepackage{bbm}
\usepackage{url}

\title{Dynamically-Scaled Deep Canonical Correlation Analysis}

\author{
Tomer Friedlander$^1$
\and
Lior Wolf$^2$
\affiliations
$^1$The School of Electrical Engineering, Tel Aviv University\\
$^2$The Blavatnik School of Computer Science, Tel Aviv University
}
\begin{document}

\maketitle

\begin{abstract}
Canonical Correlation Analysis (CCA) is a method for feature extraction of two views by finding maximally correlated linear projections of them. Several variants of CCA have been introduced in the literature, in particular, variants based on deep neural networks for learning highly correlated nonlinear transformations of two views. As these models are parameterized conventionally, their learnable parameters remain independent of the inputs after the training process, which may limit their capacity for learning highly correlated representations. We introduce a novel dynamic scaling method for training an input-dependent canonical correlation model. In our deep-CCA models, the parameters of the last layer are scaled by a second neural network that is conditioned on the model's input, resulting in a parameterization that is dependent on the input samples. We evaluate our model on multiple datasets and demonstrate that the learned representations are more correlated in comparison to the conventionally-parameterized CCA-based models and also obtain preferable retrieval results. Our code is available at \url{https://github.com/tomerfr/DynamicallyScaledDeepCCA}.
\end{abstract}

\section{Introduction}

Given two domains, the goal of CCA methods \cite{vinod1976canonical,andrew2013deep} is to recover highly correlated projections between them. The output of such methods is pairs of 1D projections, where each pair contains a single projection of each domain. Collectively, all projections from a single domain are uncorrelated, similarly to PCA.

{All existing CCA models are parameterized by conventional static parameters, i.e. their learnable parameters remain independent of the inputs after being optimized in the training process. In this work we propose to apply dynamic scaling, i.e. scaling the parameterization of the feature extractors based on the specific inputs. Such dynamic scaling is able to increase the expressiveness of the model in a way that adjusts the learned representations specifically to the inputted paired views, resulting in more correlated representations.} Importantly, to remain faithful to the CCA line of work, the representation of a vector in the first domain cannot be dependent on the input of the second domain. Each projection (linear or non-linear) has to be performed independently of the other domain. Otherwise, the settings of the projection changes, and the complexity of performing, for example, retrieval, become quadratic in the number of samples. 

The usage of dynamic scaling is shown to lead to favorable results in comparison to existing CCA methods, both classical \cite{vinod1976canonical,lopez2014randomized,michaeli2016nonparametric}
and modern deep CCA approaches \cite{andrew2013deep,wang2015deep,chang2018scalable,wang2019efficient,lindenbaum2020ell_0}. 
Specifically, we observed an increase in total correlation scores across the standard benchmarks of the field. In addition, we were able to surpass the performance of the state-of-the-art CCA-based retrieval model, Ranking CCA~\cite{dorfer2018end}, on the task of image to text retrieval.

\section{Related Work}
Canonical Correlation Analysis (CCA) \cite{harold1936relations} is a method for finding linear projections of two views, such that the projections are maximally correlated, while the projections within each view are constrained to be uncorrelated. 

CCA extensions include regularized CCA \cite{vinod1976canonical}, which employs ridge regression, Kernel CCA (KCCA) \cite{akaho,melzer2001nonlinear,bach2002kernel,hardoon2004canonical} that applies non-linear transformations using kernel functions, and scalable KCCA versions such as FKCCA and NKCCA\cite{lopez2014randomized}, which employ random Fourier features and the Nystr\"om approximation, respectively. Deep CCA (DCCA) by \cite{andrew2013deep} models the transformation functions using deep neural networks. CorrNet is an encoder-decoder architecture for maximizing the correlation between the projections of two views~\cite{chandar2016correlational}, but it does not compute canonical components. Deep canonically-correlated Autoencoder (DCCAE) \cite{wang2015deep} was proposed in parallel as an encoder-decoder model that optimizes the CCA formulation together with reconstructing the input views. NCCA \cite{michaeli2016nonparametric} is a non-parametric CCA model, which was demonstrated to match the performance of DCCA on some datasets without using neural networks.

Soft-CCA \cite{chang2018scalable} replaces the hard decorrelation constraint of the DCCA formulation with a softer constraint. It was demonstrated to be more efficient and scalable than the previous CCA-based variants. Soft-HGR \cite{wang2019efficient} is a neural framework for optimizing a softer formulation of the Hirschfeld-Gebelein-Re\'onyi (HGR) maximal correlation \cite{hirschfeld1935connection,gebelein1941statistische,renyi1959measures}.
HGR is a statistical measure of dependence, which generalizes Pearson's correlation. CCA can be regarded as a realization of HGR by a linear projection. 
Recently, $\ell_{0}$-CCA \cite{lindenbaum2020ell_0} was introduced as a sparse variant of DCCA. Sparsity is obtained by training stochastic gates, which multiply the input views. The gates are static (independent of the inputs) and multiply the inputs and not the parameters of the model.

\noindent{\bf CCA-based Cross-Modality Retrieval\quad}In the task of cross-modality retrieval, there is a source sample of one view and a set of target samples of another view. The goal of this task is to find the matching target sample of the source sample out of the set of all target samples. While retrieval can be successful even if it relies on capturing limited aspects of the data, CCA-based retrieval methods strive to construct a shared embedding space, which captures the maximal correlation of the two views. Deep CCA was used in \cite{Yan_2015_CVPR}. Later, \cite{dorfer2018end} proposed Ranking-CCA as an end-to-end CCA-based model that explicitly minimizes the pairwise ranking loss for retrieval and achieves improved retrieval results.

\noindent{\bf Dynamic Networks\quad} Neural network layers are usually parameterized by conventional static learnable parameters, which remain independent of the inputs after the training process. Dynamic Networks \cite{han2021dynamic} are neural networks, which can adapt their parameters or architecture to the inputs. For example, an adaptive ensemble of parameters was obtained by training a soft attention mechanism \cite{yang2019condconv} on multiple convolutional kernels, with attention coefficients being dependent on the inputs. Another approach is weight generation, as done by Hypernetworks \cite{ha2016hypernetworks}, which uses one neural network to generate the weights of another network.

The existing CCA models are parameterized by conventional static parameters and, at inference time, are independent of the inputs. Inspired by dynamic networks, we propose a dynamically-scaled deep CCA model. The last layer of the feature extractor of each view is modeled by a Dynamically-Scaled Layer (DSL). For each DSL, we train another neural network to generate scaling factors, which are then used to dynamically scale the conventional static parameters of the DSL, given the input view. The resulting Dynamically-Scaled Deep CCA is modelled by input-dependent transformation functions, whose improved representation power can lead to learning more correlated projections.

\section{Method}

We first present the dynamically scaled layer, and then discuss its use in Deep-CCA and Ranking-CCA models.

\subsection{Dynamically-Scaled Layer}\label{subsec:dsl}
\begin{figure}[t]
\centering
\includegraphics[width=0.99\columnwidth]{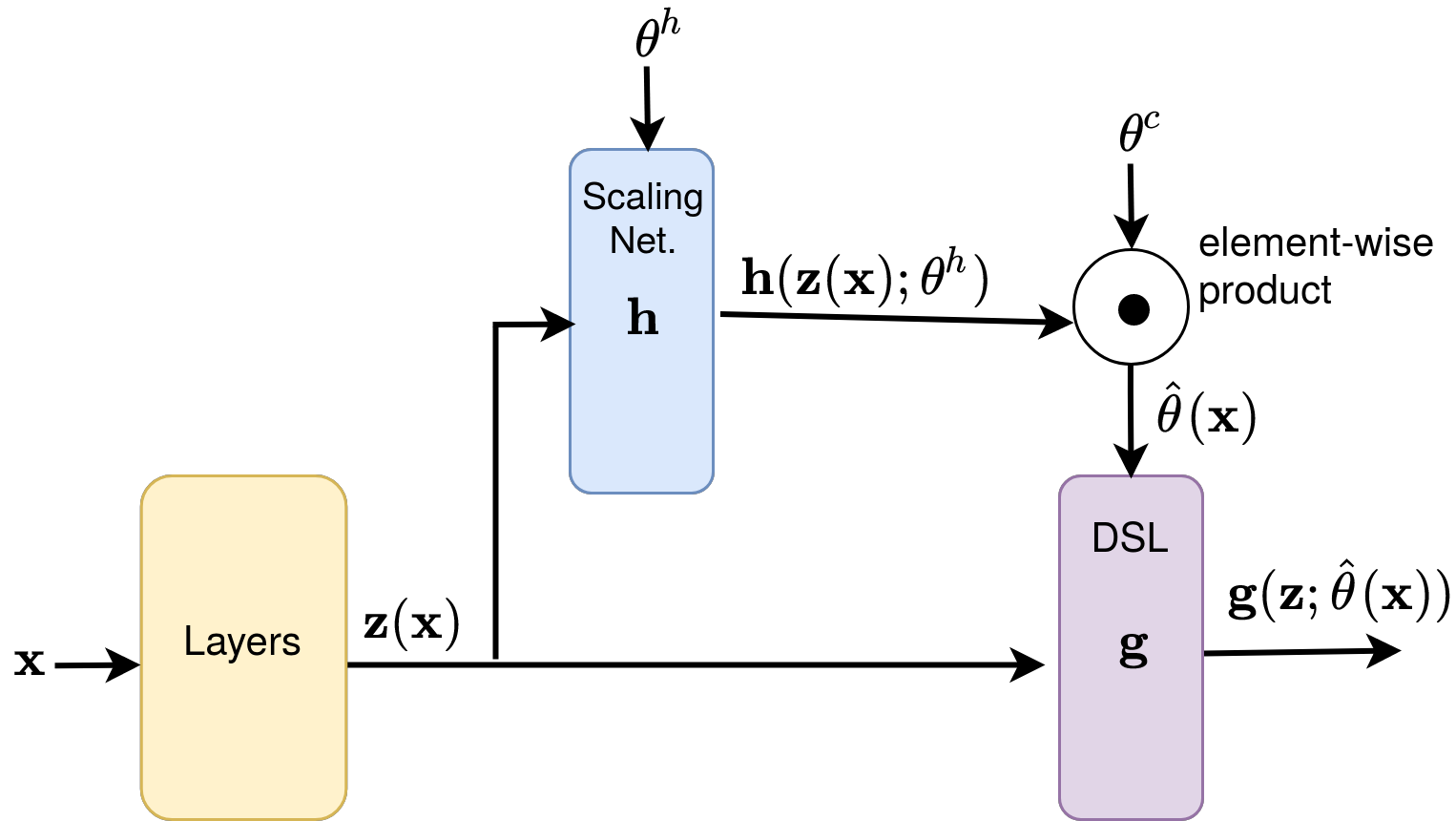}
\caption{Schematic illustration of a DSL following other neural network layers. Input vectors and the parameterization are represented by horizontal and vertical lines, respectively.}
\label{fg:dsl_fig}
\end{figure}
For the purpose of parameterizing a layer with input-dependent parameters, we train another neural network, denoted by $\mathbf{h}$. The outputs of this neural network are used as scaling factors for dynamically adjusting the parameters of the conventional layer. We name such a layer a Dynamically-Scaled Layer (DSL). A schematic illutration of a DSL within a neural network is depicted in Figure \ref{fg:dsl_fig}.

Suppose a given layer of a neural network is a DSL and it is modelled by the transformation function $\mathbf{g}$. Denote the parameters of this layer after applying the dynamic scaling by $\hat{\bm{\theta}}$. These parameters are obtained by scaling the conventional parameters of the layer, $\bm{\theta}^{c}$, with the outputs of the scaling network, $\mathbf{h}$.

The scaling network $\mathbf{h}$ is formally defined by $\mathbf{h}(\mathbf{z}(\mathbf{x});\bm{\theta}^{h})$, where $\bm{\theta}^{h}$ are the conventional parameters of the scaling network itself. $\mathbf{z}(\mathbf{x})$ is the vector of activations from the previous layer, which serves as the input of the DSL layer $\mathbf{g}$, similarly to the way information is passed between layers, and is the input of the scaling network $\mathbf{h}$. The notation of the original input of the network, denoted by $\mathbf{x}$, is explicitly used in order to emphasize that the vector $\mathbf{z}$ is a function of $\mathbf{x}$. The outputs of the scaling network, $\mathbf{h}$, are the same size as the conventional parameters, $\bm{\theta}^{c}$. 
The scaling network scales the conventional parameters according to the following equation: 
\begin{equation}
\label{eq:scaled_params}
{\hat{\bm{\theta}}}(\mathbf{x}) = \mathbf{h}(\mathbf{z}(\mathbf{x});\bm{\theta}^{h}) \odot \bm{\theta}^{c}
\end{equation}
where $\odot$ is the element-wise product operation. After applying the dynamic scaling, the parameters of $\mathbf{g}$ become input-dependent, i.e. dependent on $\mathbf{x}$. For a concrete example, let $\mathbf{g}$ be a fully-connected layer. Denote by $\mathbf{S}_{\mathbf{W}} \in \mathbb{R}^{d_{in} \times d_{out}}$ and $\mathbf{S}_{b} \in \mathbb{R}^{d_{out}}$ the outputs of the scaling network for adjusting the weight matrix, $\mathbf{W}$, and the bias vector, $\mathbf{b}$, respectively. The parameters of the fully-connected DSL are computed by the following equation:

\begin{equation}
\label{eq:scaled_fully_connected}
{\hat{\bm{\theta}}}(\mathbf{x}) = (\hat{\mathbf{W}}, \hat{\mathbf{b}}) = (\mathbf{S}_{\mathbf{W}} \odot \mathbf{W}, \mathbf{S}_{\mathbf{b}} \odot \mathbf{b})
\end{equation}
The limitation of the DSL is the increased number of parameters and the training time due to the scaling network as a function of the dimension of $\mathbf{z}$ and the total number of parameters to be scaled.

\noindent{\bf Optimization\quad}The optimization process consists of two phases. Firstly, the warm-up period is performed and the model is optimized similarly to a conventionally-parameterized neural network, i.e. the scaling network is not included in the model and ${\hat{\bm{\theta}}}(\mathbf{x}) = \bm{\theta}^{c}$ is used instead of Eq.~\ref{eq:scaled_params}. The warm-up period allows to obtain a better initialization of the conventional parameters ($\bm{\theta}^{c}$) prior to optimizing the scaling network. After a predefined number of $T$ epochs, the scaling network is added to the model and its optimization starts as well. The optimization of the scaling network ($\bm{\theta}^{h}$) is done by gradient descent as for any other parameter since the scaling network is a differentiable neural network and the scaling operator is the product operator.

\subsection{Dynamically-Scaled Deep CCA}
We introduce a novel Deep CCA model, whose transformation functions are parameterized by input-dependent parameters. Let $\mathbf{x}_{1} \in \mathbb{R}^{n_1}$ and $\mathbf{x}_{2} \in \mathbb{R}^{n_2}$ be random vectors of two views. The $d$-dimensional projections of the first and second views are computed by the transformation functions $\mathbf{f}^{*}_{1}: \mathbb{R}^{n_1} \rightarrow \mathbb{R}^{d}$ and $\mathbf{f}^{*}_{2}: \mathbb{R}^{n_2} \rightarrow \mathbb{R}^{d}$, respectively. 
Each transformation function $\mathbf{f}^{*}_{j}$ consists of a neural network feature extractor ($\mathbf{f}_{j}$) followed by a projection matrix ($\mathbf{A}_{j}$), as will be detailed next. In our method, the last layer of each feature extractor is modelled by a DSL, making them parameterized by input-dependent parameters.

Formally, denote the mapping function represented by the first stage of the feature extractor $\mathbf{f}_{j}$, i.e. without the final DSL, by $\tilde{\mathbf{f}}_{j}$ and its conventional static parameters by $\tilde{\bm{\theta}}_{j}^{c}$. The output of the first stage of the feature extractor of the $j^{th}$ view for the input $\mathbf{x}_{j}$ is denoted by $\mathbf{z}_{j}(\mathbf{x}_{j})=\tilde{\mathbf{f}_{j}}(\mathbf{x}_{j};\tilde{\bm{\theta}}_{j}^{c})$.

The DSL scaling networks $\mathbf{h}_{1}$ and $\mathbf{h}_{2}$ with the conventional static parameters $\bm{\theta}_{1}^{h}$ and $\bm{\theta}_{2}^{h}$ are defined for the first and second views, respectively. In addition, the mapping function of the DSL of $\mathbf{f}_{j}$ and its conventional static parameters are denoted by $\mathbf{g}_{j}$ and $\bm{\theta}_{j}^{c}$, respectively. The scaling network $\mathbf{h}_{j}$ scales the corresponding conventional parameters $\bm{\theta}_{j}^{c}$ according to Equation \ref{eq:scaled_params}, resulting in the scaled parameters $\hat{\bm{\theta}}_{j}(\mathbf{x}_{j})$. In total, the extracted features of the $j^{th}$ view are computed by $\mathbf{f}_{j}(\mathbf{x}_{j};\hat{\bm{\Theta}}_{j}({\mathbf{x}_{j}}))=\mathbf{g}_{j}\big(\mathbf{z}_{j};\hat{\bm{\theta}}_{j}(\mathbf{x}_{j})\big)$, where the total parameterization is denoted by $\hat{\bm{\Theta}}_{j}(\mathbf{x}_{j})=\big\{\tilde{\bm{\theta}}_{j}^{c},\hat{\bm{\theta}}_{j}(\mathbf{x}_{j})\big\}$ and $\mathbf{z}_{j}$ was omitted for a convenient notation. Finally, the total projection of the $j^{th}$ view to the shared space is computed by projecting the extracted features by $\mathbf{A}_{j}$ as follows:
\begin{equation}
\label{eq:dcca_projection}
\begin{aligned}
\mathbf{f}_{j}^{*}(\mathbf{x}_{j};\hat{\bm{\Theta}}_{j}(\mathbf{x}_{j}))={\mathbf{A}^{\top}_{j}}\mathbf{f}_{j}(\mathbf{x}_{j};\hat{\bm{\Theta}}_{j}(\mathbf{x}_{j}))
\end{aligned}
\end{equation}
{We present following extension of the Deep CCA optimization problem, which includes dynamically-scaled transformation functions:}
\begin{equation}
\label{eq:dsdcca_formulation}
\begin{aligned}
&\argmax_{\{\bm{\Theta}_{j}^{c},\bm{\theta}_{j}^{h},\mathbf{A}_{j}\}_{j=1}^{2}}  tr\Bigg(cov\Big(\mathbf{f}^{*}_{1}(\mathbf{x}_{1};\hat{\bm{\Theta}}_{1}(\mathbf{x}_{1})),\mathbf{f}^{*}_{2}(\mathbf{x}_{2};\hat{\bm{\Theta}}_{2}(\mathbf{x}_{2}))\Big)\Bigg) \\
&\textrm{s.t.} \\
&\qquad \qquad cov\Big(\mathbf{f}^{*}_{1}\big(\mathbf{x}_{1};\hat{\bm{\Theta}}_{1}(\mathbf{x}_{1})\big),\mathbf{f}^{*}_{1}\big(\mathbf{x}_{1};\hat{\bm{\Theta}}_{1}(\mathbf{x}_{1})\big)\Big)=\mathbf{I} \\
&\qquad \qquad cov\Big(\mathbf{f}^{*}_{2}\big(\mathbf{x}_{2};\hat{\bm{\Theta}}_{2}(\mathbf{x}_{2})\big),\mathbf{f}^{*}_{2}\big(\mathbf{x}_{2};\hat{\bm{\Theta}}_{2}(\mathbf{x}_{2})\big)\Big)=\mathbf{I} \\
\end{aligned}
\end{equation}

where $tr$ is the trace operator and $cov$ is the computation of the covariance matrix. The conventional static parameters of the entire $\mathbf{f}_{j}$ network, without the parameters of the scaling network, are denoted by $\bm{\Theta}_{j}^{c}=\{\tilde{\bm{\theta}}_{j}^{c},\bm{\theta}_{j}^{c}\}$. 

\noindent{\bf Optimization\quad}{For training the Dynamically-Scaled Deep CCA, we follow the optimization process of the original Deep CCA \cite{andrew2013deep}, but with our extended formulation of dynamically scaled transformation functions, as described next. Suppose a set of paired instances of the two views is given, i.e. $\{(\mathbf{x}_{1}^{i},\mathbf{x}_{2}^{i})\}_{i=1}^{N}$, where $\mathbf{x}_{1}^{i} \in \mathbb{R}^{n_1}$ and $\mathbf{x}_{2}^{i} \in \mathbb{R}^{n_2}$. The obtained extracted features of the views are denoted by $\{\mathbf{f}_{1}(\mathbf{x}_{1}^{i};\hat{\bm{\Theta}}_{1}(\mathbf{x}_{1}^{i}))\}_{i=1}^{N}$ and $\{\mathbf{f}_{2}(\mathbf{x}_{2}^{i};\hat{\bm{\Theta}}_{2}(\mathbf{x}_{2}^{i}))\}_{i=1}^{N}$, respectively. These projections are normalized to have a zero mean with respect to averaging across the sample index $i$, i.e. $\bar{\mathbf{f}}_{j}(\mathbf{x}_{j}^{i};\hat{\bm{\Theta}}_{j}(\mathbf{x}_{j}^{i}))=\mathbf{f}_{j}(\mathbf{x}_{j}^{i};\hat{\bm{\Theta}}_{j}(\mathbf{x}_{j}^{i}))-\frac{1}{N}\sum_{n=1}^{N}{\mathbf{f}_{j}(\mathbf{x}_{j}^{n};\hat{\bm{\Theta}}_{j}(\mathbf{x}_{j}^{n}))}$. Let $\mathbf{\bar{F}}_{1} \in \mathbb{R}^{d \times N}$ and $\mathbf{\bar{F}}_{2} \in \mathbb{R}^{d \times N}$ be matrices, whose columns are $\{\bar{\mathbf{f}}_{1}(\mathbf{x}_{1}^{i};\hat{\bm{\Theta}}_{1}(\mathbf{x}_{1}^{i}))\}_{i=1}^{N}$ and $\{\bar{\mathbf{f}}_{2}(\mathbf{x}_{2}^{i};\hat{\bm{\Theta}}_{2}(\mathbf{x}_{2}^{i}))\}_{i=1}^{N}$ respectively.

The estimated cross-covariance matrix of the extracted features of each view is computed by $\bm{\Sigma}_{1,2}=\frac{1}{N-1}\bar{\mathbf{F}}_{1}\bar{\mathbf{F}}_{2}^{T}$. Similarly, define the estimated auto-covariance matrices of each view by $\bm{\Sigma}_{1,1}=\frac{1}{N-1}\bar{\mathbf{F}}_{1}\bar{\mathbf{F}}_{1}^{T}+r_{1}\mathbf{I}$ and $\bm{\Sigma}_{2,2}=\frac{1}{N-1}\bar{\mathbf{F}}_{2}\bar{\mathbf{F}}_{2}^{T}+r_{2}\mathbf{I}$, where $r_{1} > 0$ and $r_{2} > 0$ are fixed hyperparameters used to ensure that $\bm{\Sigma}_{1,1}$ and $\bm{\Sigma}_{2,2}$ are positive definite. The optimization problem from Equation \ref{eq:dsdcca_formulation} in terms of the estimated covariance matrices becomes as follows:

\begin{equation}
\label{eq:approx_dsdcca_formulation}
\begin{aligned}
&\argmax_{\{\bm{\Theta}_{j}^{c},\bm{\theta}_{j}^{h},\mathbf{A}_{j}\}^{2}_{j=1}}  tr(\mathbf{A}^{\top}_{1}\bm{\Sigma}_{1,2}\mathbf{A}_{2}) \\
&\textrm{s.t.} \quad \mathbf{A}^{\top}_{1}\bm{\Sigma}_{1,1}\mathbf{A}_{1}=\mathbf{I} ,\quad \mathbf{A}^{\top}_{2}\bm{\Sigma}_{2,2}\mathbf{A}_{2}=\mathbf{I} \\
\end{aligned}
\end{equation}

Let $\bm{\Psi}=\bm{\Sigma}_{1,1}^{-1/2}\bm{\Sigma}_{1,2}\bm{\Sigma}_{2,2}^{-1/2}$ and let $\{\epsilon_k\}_{k=1}^{d}$ be the set of the top $d$ singular values of the matrix $\bm{\Psi}$. In order to find the optimal parameters of the feature extractors in Equation \ref{eq:approx_dsdcca_formulation}, we minimize the same loss used in the original Deep CCA formulation \cite{andrew2013deep}, given by 
\begin{equation}
\label{eq:dcca_loss}
\mathcal{L}_{dcca} = -\sum_{k=1}^{d}{\epsilon_{k}} 
\end{equation}

The gradients for minimizing the Deep CCA loss are detailed in \cite{andrew2013deep}. As demonstrated by \cite{wang2015unsupervised,wang2015deep}, the Deep CCA loss can be optimized effectively by a gradient descent optimizer on sufficiently-large mini-batches. The two-stage optimization scheme from Sec. \ref{subsec:dsl} is applied.

The optimal projection matrices, $\mathbf{A}_{1}$ and $\mathbf{A}_{2}$, are computed only after optimizing the feature extractors, by $\mathbf{A}_{1}=\bm{\Sigma}^{-1/2}_{1,1}\mathbf{U}$ and $\mathbf{A}_{2}=\bm{\Sigma}^{-1/2}_{2,2}\mathbf{V}$, where the matrices $\mathbf{U}$ and $\mathbf{V}$ consist of the left and right singular vectors of $\bm{\Psi}$, which correspond to $\{\epsilon_{k}\}_{k=1}^{d}$. 
}

\subsection{Dynamically-Scaled Ranking CCA}
We present another usage of dynamic scaling for the purpose of improving a CCA-based model. In particular, we present a dynamically-scaled variant of the Ranking-CCA model \cite{dorfer2018end}, which was specifically designed for the cross-modal retrieval task.

A successful cross-modality retrieval model has to be able to distinguish well enough between mismatched samples. For this purpose, we propose to input the scaling networks of the DSL with the concatenation of both $\mathbf{z}$ and the original raw input to the network, $\mathbf{x}$. The additional input is able to add more context for learning how to dynamically scale the parameters, such that the resulting embedding vectors are distinguishable. In particular, $\mathbf{h}(\mathbf{z}(\mathbf{x});\bm{\theta}^{h})$ in Equation \ref{eq:scaled_params} is replaced by $\mathbf{h}([\mathbf{z}(\mathbf{x}),\mathbf{x}];\bm{\theta}^{h})$. Let $\mathbf{f}^{*}_{1}$ and $\mathbf{f}^{*}_{2}$ be the dynamically-scaled transformation functions for the first and second views, respectively, as denoted in Eq. \ref{eq:dcca_projection}.

Following the original Ranking-CCA, the feature extractors and the projection matrices are not trained separately as done in Deep CCA, but they are trained concurrently at each training iteration. In particular, for each mini-batch of training samples, the projection matrices are explicitly computed as described above. In addition, instead of minimizing the Deep CCA loss $\mathcal{L}_{dcca}$ for optimizing the feature extractors, \cite{dorfer2018end} suggests to optimize the entire architecture by end-to-end minimization of the pairwise ranking loss. This loss is tailored for the cross-modality retrieval task, by encouraging matching samples to be closer in the embedding space than mismatching samples. Each projection matrix serves as a differnetiable linear layer on top of the feature extractors and allows backpropagation of the gradients of the optimized loss to the previous layers. The pairwise ranking loss is computed as follows:
\begin{equation}
\label{eq:pairwise_ranking_loss}
\begin{aligned}
\sum_{i,j\neq i}\mathbbm{1}\Big(m - s\big(\mathbf{f}_{1}^{*}(\mathbf{x}_{1}^{i}),\mathbf{f}_{2}^{*}(\mathbf{x}_{2}^{i})\big) + s\big(\mathbf{f}_{1}^{*}(\mathbf{x}_{1}^{i}),\mathbf{f}_{2}^{*}(\mathbf{x}_{2}^{j})\big)\Big)
\end{aligned}
\end{equation}
where s is a scoring function of two input vectors, e.g. the cosine similarity, and $\mathbbm{1}(u)=max(0,u)$. The hyperparameter $m$ is the margin. Summing over $j$ is done across all mismatching samples of $\mathbf{x}_{1}^{i}$ within the batch. A symmetric pairwise loss, is defined by switching each notation of 1 by 2 in Equation \ref{eq:pairwise_ranking_loss}, and vice versa. The parameterization is omitted in Eq. \ref{eq:pairwise_ranking_loss} in order to simplify the notations, but both $\mathbf{f}_{1}^{*}$ and $\mathbf{f}_{2}^{*}$ are dynamically-scaled in contrast to the original formulation. 

\noindent{\bf Cross-Modality Retrieval\quad} Assume we are given a source instance of the first view, $\mathbf{x}_{1}\in\mathbb{R}^{n_1}$, and a set of target instances of the second view, $\{\mathbf{x}_{2}^{i}\}$, where $\mathbf{x}^{i}_{2} \in \mathbb{R}^{n_{2}}$. The projections of $\mathbf{x}_{1}$ and all target set instances are computed, resulting in $\mathbf{f}_{1}^{*}(\mathbf{x}_{1};\hat{\bm{\Theta}}_{1}(\mathbf{x}_{1}))$ and $\{\mathbf{f}_{2}^{*}(\mathbf{x}_{2}^{i};\hat{\bm{\Theta}}_{2}(\mathbf{x}_{2}^{i}))\}_{i}$. The $k$ closest projected target samples to the projected source sample in terms of the cosine-distance, are chosen as the top-$k$ suggestions by the model.

\section{Experiments}

In order to evaluate the contribution of dynamic scaling, we perform two sets of experiments. One is focused on the total canonical correlation score, which is often used in the CCA literature. The second set focuses on the retrieval application.

The neural networks consist of fully-connected layers, whose number and width follow the literature (supplementary). For a given dataset, we train all neural network-based transformation functions with the same architecture for the same number of training epochs, after being initialized by the same seed. After each training epoch, the loss of the current checkpoint of the model is computed. The learnable parameters that achieve the lowest loss on the validation set are chosen as the parameters of the model. The hyperparameters of each model are tuned by evaluating, on the validation set, the total canonical correlation and recall measurements for the total canonical correlation and retrieval experiments, respectively. Regarding our proposed DS-DCCA and DS-Ranking CCA, we model the conventionally static layers of the mapping functions with neural networks, which are identical to the networks modelling the mapping functions in the network-based baselines. The scaling networks are fully connected layers followed by a batch normalization \cite{ioffe2015batch} and a ReLU \cite{nair2010rectified} activation function. See supplementary material for the full implementation details.

\begin{table*}
\centering
\begin{tabular}{l|rr|rr|rr} 
\toprule
\multirow{2}{*}{Model} & \multicolumn{2}{c|}{MNIST} & \multicolumn{2}{c|}{XRMB} & \multicolumn{2}{c}{Flickr8k}  \\ 
\cline{2-7}
& MEAN$\pm$STD & P-VALUE & MEAN$\pm$STD & P-VALUE & MEAN$\pm$STD & P-VALUE     \\ 
\hline
Upper Bound ($d$) &50.00$\pm$0.00  &-  &112.00$\pm$0.00  &-  &128.00$\pm$0.00  &- \\
\hline
CCA &28.88$\pm$0.27  &1.76E-12  &15.88$\pm$0.07  &3.34E-13  &41.58$\pm$0.61  &2.07E-13 \\
FKCCA &41.71$\pm$0.07 &1.78E-12  &95.45$\pm$0.18  &1.04E-09  &39.24$\pm$0.60  &7.31E-13 \\
NKCCA &45.10$\pm$0.04  &6.45E-11  &103.49$\pm$0.14  &1.10E-08  &68.49$\pm$1.00  &4.31E-11 \\
DCCA &46.76$\pm$0.03  &2.18E-08  &108.73$\pm$0.10  &1.01E-07  &67.65$\pm$1.28  &2.60E-10 \\
DCCAE &46.75$\pm$0.02  &1.07E-08  &108.71$\pm$0.09  &3.17E-08  &67.75$\pm$1.18  &1.29E-10 \\
NCCA &40.98$\pm$0.12  &1.37E-11  &107.57$\pm$0.18  &5.11E-07  &57.47$\pm$4.07  &1.27E-07 \\
Soft-CCA &44.55$\pm$0.18  &1.38E-08  &84.85$\pm$1.43  &2.50E-06  &60.55$\pm$1.05  &5.92E-11 \\
Soft-HGR &46.86$\pm$0.04  &1.54E-07  &106.12$\pm$0.11  &5.68E-08  &54.41$\pm$4.35  &1.42E-07 \\
$\ell_{0}$-CCA &47.17$\pm$0.03  &3.13E-06  &108.72$\pm$0.11  &1.99E-07  &72.83$\pm$1.11  &1.08E-09 \\
DS-DCCA &\textbf{47.50$\pm$0.03}  &-  &\textbf{110.88$\pm$0.06}  &-  &\textbf{86.06$\pm$1.28}  &- \\
\bottomrule
\end{tabular}
\caption{Total correlation of top $d$ canonical components. Means and standard deviations of the test sets obtained in k-folds cross validation are reported, as well as the p-value computed via a paired t-test vs. our DS-DCCA.}
\label{tbl:total_correlation}
\end{table*}

\begin{table}
\centering
\begin{tabular}{l|rr|rr|rr} 
\toprule
\multirow{2}{*}{Model} & \multicolumn{2}{c|}{MNIST } & \multicolumn{2}{c|}{XRMB } & \multicolumn{2}{c}{Flickr8k}  \\ 
\cline{2-7}
                       & TC & \%                     & TC & \%                    & TC & \%                       \\ 
\hline

Up. Bound  &50.00    &{-}     &112.00    &{-}      &128.00    &{-}                \\ 
\midrule
\textbf{\underline{DCCA}}:  &    &     &    &      &    &   \\ 
Original  &46.71    &26     &108.69    &65      &67.51    &30               \\ 
Wide 2  &46.90    &21     &109.67    &50      &76.74    &17         \\ 
Wide 1\&2  &46.86    &22     &109.29    &57      &76.15    &18                 \\ 
Global Scale &46.70    &26     &108.63    &66      &67.65    &30                 \\ 
\midrule
\textbf{\underline{DS-DCCA}}: &    &   &    &    &    &  \\
Scaling outputs & 47.22 & 12 & 110.12 & 39 & 79.11 & 13 \\
w/o $\bm{\theta}^{c}$  &47.42    &5     &110.78    &5      &80.25    &11              \\ 
No warm-up   &47.36    &8     &110.47    &24      &80.65    &10                 \\ 
Full config.  &\textbf{47.56}    &{-}     &\textbf{110.84}    &{-}      &\textbf{85.57}    &{-}                 \\ 

\bottomrule
\end{tabular}
\caption{Ablation study - The total correlation (TC) is reported as well as the percent of the remaining gap to the upper bound, which is improved by the full configuration of our DS-DCCA.} 
\label{tbl:ablation}
\end{table}

\subsection{Total Canonical Correlation} 
We first evaluate our Dynamically-Scaled Deep CCA (DS-DCCA) model with respect to the total canonical correlation score on the common benchmarks of the CCA literature and compare it with several CCA-based models. These experiments directly test the performance of the model with respect to the main objective of the CCA formulation. 

\noindent{\bf Baselines\quad} We compare our proposed DS-DCCA model to well-known CCA-based models from the literature. In particular, methods modelled by classical approaches with no neural networks: (1) regularized CCA \cite{vinod1976canonical}, (2) FKCCA \cite{lopez2014randomized}, (3) NKCCA \cite{lopez2014randomized}, (4) NCCA \cite{michaeli2016nonparametric}. In addition, we compare our method to CCA methods, which are modelled by neural-networks: (5) DCCA \cite{andrew2013deep}, (6) DCCAE \cite{wang2015deep}, (7) Soft-CCA \cite{chang2018scalable}, (8) Soft-HGR \cite{wang2019efficient}, (9) $\ell_{0}$-CCA \cite{lindenbaum2020ell_0}. 

\noindent{\bf Datasets\quad}Three datasets were used for evaluation: MNIST, {Wisconsin X-Ray Microbeam (XRMB)}, and Flickr8k. {{\textbf{MNIST}}} \cite{lecun1998gradient} consists of $28\times28$ grayscale images of handwritten digits. We employ a variant of MNIST that is used in the CCA literature \cite{chandar2016correlational,chang2018scalable}, which defines the two different views of a given image to be its left and right halves, respectively. \cite{chandar2016correlational} splits the dataset to 50k/10k/10k for training/validating/testing. The dimension of the projections is selected to be $d=50$ as used in the CCA literature. {\textbf XRMB} \cite{westbury1994x} consists of simultaneously recorded speech and articulatory measurements. The acoustic features are Mel-frequency cepstral coefficients, which yield a 112-dimensional vector. The articulatory measurements are displacements of the speaker's vocal tract, which yield a 273-dimensional vector. \cite{lopez2014randomized} splits the dataset to 30k/10k/10k for training/validating/testing. The dimension of the projections is selected to be $d=112$ as used in the CCA literature. 

{\textbf{Flickr8k}} \cite{hodosh2013framing} is a dataset consisting of 8,000 images from Flickr.com and five textual captions per each image. We encode each image to a 2048d embedding vector by the penultimate layer of a ResNet50 neural network \cite{he2016deep}, which was pre-trained on ImageNet \cite{deng2009imagenet}. After removing stop-words and applying the SpaCy lemmatization \cite{spacy}, each set of five captions is embedded to a single 300d vector by mean-pooling the Word2Vec \cite{word2vec} embedding vectors.

\noindent{\bf Total canonical correlation results\quad}The results for the total correlation scores are presented in Table \ref{tbl:total_correlation}. Each model is trained on the training set to learn a $d$-dimensional representation of each view. Following prior works \cite{andrew2013deep,michaeli2016nonparametric}, another regularized linear CCA \cite{vinod1976canonical} is trained on the projected training samples of each view, which ensures that the top $d$ canonical components of each representation are extracted. The hyperparameters of each model are selected to be those that achieve the best total canonical correlation on the validation set. The total correlation is measured by summing the correlation coefficients between each pair of corresponding canonical components. Each model with its best perfomring hyperparameters is then evaluated using k-fold cross validation, which preserves the above mentioned subset ratios, i.e. 7, 5 and 8 folds for MNIST, XRMB and Flickr8k, respectively. For example, for each testing fold out of the 7 folds for MNIST, the remaining 6 folds are randomly split to 50k and 10k samples for training and validating.

For each model and each dataset, we report in Table \ref{tbl:total_correlation} the mean and standard deviation of the model's results on the test set after performing k-fold cross validation. In addition, we report for each baseline the p-values computed via a paired t-test vs. our DS-DCCA. The mentioned upper bound of the total canonical correlation for each case is equal to the total number of canonical components ($d$).

Evidently, the proposed DS-DCCA learns canonical components of the input views, which are more correlated on average in comparison to both classical and modern CCA-based models on the compared benchmarks. In particular, DS-DCCA improves the second best result by 11.7\%, 65.9\% and 24.0\% of the remaining gap to the upper bound for MNIST, XRMB and Flickr8k, respectively. Moreover, the very low p-values emphasize that the leading results of our DS-DCCA are statistically significant.

\noindent{\bf Ablation\quad}We perform an ablation study on the total correlation experiment for the three mentioned datasets: MNIST, XRMB and Flickr8k. In particular, we compare our proposed DS-DCCA model (full config.) to: (1) The original variant of DCCA, (2) DCCA with a wider middle layer and (3) DCCA with wider first and second layers, reaching a comparable number of added parameters as the scaling networks, (4) DCCA, whose parameters of the last layer for each view are scaled by learnable scaling factors, which are independent of the inputs and remain static after the training (DCCA + Global Scaling). (5) A dynamically-scaled DCCA with a similar capacity to our proposed model, but its scaling networks scale the outputs of the last layer instead of its parameters (Scaling outputs), (6) DCCA model, whose last layer is parameterized by a hypernetwork, i.e. this layer is not parameterized by conventional parameters ($\bm{\theta}^{c}$) scaled by another network. Instead, the parameters are generated directly by the outputs of another network for the input $\mathbf{z}$ (w/o $\bm{\theta}^{c}$). (6) DS-DCCA, whose scaling networks were trained from the beginning of the process without a warm-up period (No warm-up).

The results are provided in Table \ref{tbl:ablation}. For each scenario, we report the total correlation, obtained on a single test set, and the percent of the remaining gap to the upper bound, which is improved by the full configuration of our DS-DCCA. As can be seen, the full configuration of our model, outperforms the compared variants. Our added scaling networks outperform the addition of neurons to each of the first layers of DCCA for reaching a comparable number of parameters. The added neurons have static and input-independent parameters, in contrast to our dynamically-scaled ones. The global scacling struggles to improve the original DCCA and even worsens the results on some datasets. These learnable scaling factors, which are independent of the inputs, are found not beneficial and make the training process more difficult. In addition, our approach of dynamically scaling the parameters of the transformation functions is superior than dynamically scaling only their outputs. We note that scaling the parameters provide many more degress of freedom than scaling the output vector. It is also evident that the warmup-period improves the results since it allows to initialize the conventional static parameters better, prior to optimizing the scaling networks. The lower result of the full hypernetwork model (no $\bm{\theta}^{c}$) also indicates that the scaling requires a network with weights that are not dynamic.

\begin{figure}[t]
\centering
\begin{tabular}{@{}c@{}}
\includegraphics[width=0.78495\linewidth]{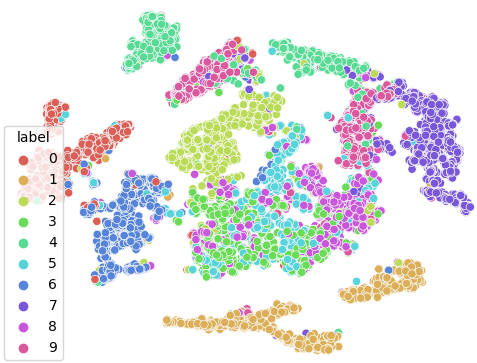}\\
\hline
    \includegraphics[width=0.78495\linewidth]{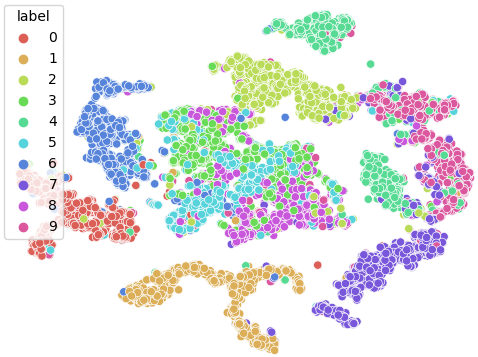}
\end{tabular}
\caption[t-SNE plots of the scaling networks' outputs for MNIST]{t-SNE plots of the scaling network outputs for the left and right halves of MNIST. The scales display a per-class behavior despite training without labels.}
\label{fig:tsne}
\end{figure}

\noindent{\bf Scales Visualization\quad}
For the purpose of visualizing the scales generated by the scaling networks ($\mathbf{h}_{1}$ and $\mathbf{h}_{2}$), we provide in Figure \ref{fig:tsne} the t-SNE plots \cite{van2008visualizing} of the scaling networks' outputs for the two views (left and right halves) of MNIST's test set. The obtained outputs are well clustered according to the input's digit. It is important to note that the model is not provided with the labels during training, and it is trained to match the two halves of the same digit.

The visualization in Figure \ref{fig:tsne} demonstrates that the obtained dynamic scales display a per-class behavior despite training without labels. This observation suggests that our dynamically-scaled approach would excel in scenarios in which the dataset samples were derived from (hidden) hierarchies of categories, since in such cases, the input presents class-based variability that requires the ability to adapt.

\subsection{Cross-Modality Retrieval}
\begin{table}
\centering
\begin{tabular}{l|rrr|rrr} 
\toprule
\multirow{2}{*}{Model} & \multicolumn{3}{c|}{IMG$\rightarrow$TXT}                                            & \multicolumn{3}{c}{TXT$\rightarrow$IMG}                                             \\ 
\cline{2-7}
                        & {$R_{1}$}                    & {$R_{5}$}                    & {$R_{10}$}                     & {$R_{1}$}                    & {$R_{5}$}                    & {$R_{10}$}                     \\ 

\midrule
\textbf{\underline{Flickr8k}:} &  & & & & &   \\
CCA   &27.3     &57.0 &68.3         &24.5     &55.5     &68.4   \\
FKCCA   &17.4     &42.9     &56.5         &17.3     &43.6     &55.4   \\
NKCCA   &27.0     &58.5     &73.8         &26.5     &57.8     &71.2   \\
DCCA   &31.8     &65.7     &77.7         &29.3     &62.8     &76.1   \\
DCCAE   &33.6     &65.9     &76.6         &30.0     &63.0     &76.0   \\
NCCA   &8.0     &25.5     &34.6         &8.5     &24.3     &34.7 \\
Soft-CCA   &28.8     &61.6     &73.5         &27.1     &60.1     &72.3   \\
Soft-HGR   &25.4     &55.1     &69.0         &21.5     &54.3     &68.8   \\ 
$\ell_{0}$-CCA   &30.7     &62.4     &76.7         &28.5     &61.2     &74.9   \\
R. CCA   &33.6     &66.9     &78.0         &31.9     &64.6     &78.6   \\
DS-R. CCA   &     &    &         &     &     &   \\
\quad $\mathbf{z}$ only   &33.7     &68.1    &\underline{79.9}         &33.1     &\underline{67.1}     &79.5   \\
\quad $\mathbf{x}$ only   &\underline{33.9}     &\underline{69.8}    &79.8         &\bfseries 34.7     &66.9     &\underline{80.0}   \\
\quad $\mathbf{z}$ \& $\mathbf{x}$   &\bfseries 35.3     &\bfseries 70.7     &\bfseries 82.7         &\underline{34.4}     &\bfseries 68.3     &\bfseries 82.3   \\
\midrule
\midrule
\textbf{\underline{Flickr30k}:}               &                      &                      &                       &                      &                      &                       \\
R. CCA   &40.2     &72.8     &83.2         &40.0     &70.7     &82.7   \\
DS-R. CCA   &      &     &         &     &     &    \\
\quad $\mathbf{z}$ only   &41.3      &\underline{74.2}     &82.8         &40.7     &\bfseries 73.4    &\underline{84.7}    \\
\quad $\mathbf{x}$ only   &\bfseries 44.1      &74.0     &\underline{84.1}         &\underline{43.9}     &\underline{73.1}     &\bfseries 85.4    \\
\quad $\mathbf{z}$ \& $\mathbf{x}$   &\underline{43.9}      &\bfseries 75.0     &\bfseries 84.5         &\bfseries 44.2     &\bfseries 73.4     &84.2    \\
\midrule
\midrule
\textbf{\underline{IAPR TC$_{\mathbf{12}}$}:}              &                      &                      &                       &                      &                      &                       \\

R. CCA   &48.6     &\underline{81.1}     &90.0         &49.0     &80.5     &89.1   \\
DS-R. CCA   &      &     &         &     &     &    \\
\quad $\mathbf{z}$ only   &48.9      &80.3     &\underline{90.2}         &49.5     &80.3     &89.0    \\
\quad $\mathbf{x}$ only   &\underline{49.1}      &\bfseries 82.7     &\underline{90.2}         &\bfseries 51.1     &\underline{81.3}     &\bfseries 91.1    \\
\quad $\mathbf{z}$ \& $\mathbf{x}$   &\bfseries 49.6       &\bfseries 82.7      &\bfseries 91.4         &\underline{50.0}      &\bfseries 82.0      &\underline{90.9}     \\
\bottomrule
\end{tabular}
\caption{Recall rates for Flickr8k, Flickr30k and IAPR TC-12. Bold and underlined results are first and second places, respectively.}
\label{tbl:retrieval}
\end{table}
We evaluate our dynamically-scaled variant of Ranking CCA (DS-R. CCA) in a cross-modality retrieval task on several image-text datasets. Given one sample of a modality, the goal of the cross-modality retrieval task is to retrieve the matching sample of the other modality from a set of samples. We evaluate the models on image annotation ($IMG\rightarrow TXT$), for which an image is given and the matching textual caption should be retrieved, and vice versa ($TXT\rightarrow IMG$), where a textual caption is given and the matching image is to be found. 
The performance is measured by recall statistics: the score $R_{k}$ is computed as the percent of source samples for which the model retrieved the correct target sample as one of its top $k$ suggestions from the total number of source samples.

\noindent{\bf Datasets\quad}Three datasets were used for evaluating the retrieval task: (1) Flickr8k, (2) Flickr30k and (3) IAPR TC-12. All datasets were pre-processed and encoded to embedding vectors as described for Flickr8k in the total canonical correlation experiment. The dimension of the projections of each of the two views is selected to be $d=128$ for all datasets. \textbf{Flickr8k} \cite{hodosh2013framing} was split as described in the previous section. \textbf{Flickr30k} \cite{young2014image} is an extended version of the Flickr8k dataset and consists of 31,784 images from Flickr.com. Following \cite{Yan_2015_CVPR}, random 1k and 1k samples are selected for validation and testing, respectively. \textbf{IAPR TC-12} \cite{grubinger2006iapr} contains 20k images. Each image is described by one caption, which is more detailed than the above-mentioned datasets. Following \cite{dorfer2018end}, random 2k and 1k samples are selected for validation and testing, respectively. 

\noindent{\bf Baselines\quad}For Flickr8k, the full configuration of our dynamically-scaled variant of Ranking CCA (DS-R.CCA) is compared to all CCA-based baselines from the total canonical correlation experiment. In addition, our variant is compared to the original non-dynamically scaled variant of Ranking CCA (R. CCA) \cite{dorfer2018end}, which was proposed specifically to learn representations for the purpose of the cross-modality retrieval task. For Flickr30k and IAPR TC-12, our model is compared to Ranking CCA (R. CCA). In order to demonstrate the effectiveness of inputting the concatenation of both $\mathbf{z}$ and $\mathbf{x}$ to the scaling networks, we compare this full configuration to variants of our DS-R.CCA, whose scaling networks are conditioned only on either $\mathbf{z}$ or $\mathbf{x}$.

\noindent{\bf Retrieval Results\quad}The recall results in the cross-modality retrival task on Flickr8k, Flickr30k and IAPR TC-12 are presented in Table \ref{tbl:retrieval}, {where the best performing model and the second one are in bold and underlined, respectively. } For each measurement of $R_{k}$, the reported results are the $R_{k}$ obtained on the test set by the hyperparameters, which achieved the best $R_{k}$ on the validation set.

As expected, Ranking-CCA outperforms all CCA-based models, which are not designed specifically for the retrieval task. Evidently, the full configuration of our DS-R.CCA ($\mathbf{z}$ and $\mathbf{x}$) achieves the best recall retrieval rates in 13 measurements out of 18, and achieves the second best values in 4 out of the remaining 5 measurements. The second best performing model among the compared ones is the $\mathbf{x}$-only variant of our DS-R.CCA, which emphasizes the importance of conditioning the dynamically scaled parameters on the raw inputs for learning distinguishable embedding vectors.

\section{Conclusion}
CCA methods learn orthogonal directions or, more generally, non-linear scalar mappings, in each of the views. In this work, we show the advantage of having computed weights, which vary based on the input pair, for scaling these scalar features so that matching between the two views is maximized. The method is based on a new type of layer that is applied as the last layer of the deep CCA method and that receives the activations of the previous layer. Our experiments demonstrate a clear advantage with regards to the total correlation score and, for the cross-modality retrieval task, the recall rate. 

\section{Acknowledgments}
This project has received funding from the European Research Council (ERC) under the European Unions Horizon 2020 research and innovation programme (grant ERC CoG 725974).

\bibliographystyle{named}
\bibliography{main}

\clearpage
\appendix
\section*{Supplementary Material}
We provide the implementation details of the experiments conducted in this work (see \ref{seq:implementation}). In addition, we provide additional results regarding the sensitivity of the compared models to the initial seed (see \ref{subseq:tc_multi_seeds} and \ref{subseq:retrieval_multi_seeds} for the total canonical correlation experiment and for the cross-modality retrieval task, respectively). 

\section{Implementation Details}\label{seq:implementation} Following the literature of the CCA baselines, all neural networks used in this paper are fully-connected networks. For a given dataset, we train all neural network-based transformation functions with the same architecture initialized by the same seed. Each fully connected layer is followed by a batch normalization layer \cite{ioffe2015batch} with no affine parameters and is then followed by the ReLU activation function \cite{nair2010rectified}. The last layer is followed by a batch normalization layer, but not by an activation function. The number of layers and neurons of each mapping function for each dataset is given in Table \ref{tbl:network_architectures}, where we follow \cite{wang2015stochastic} for MNIST and XRMB. The training is done for the same number of epochs (100, 500, 1k, 200 and 200 for MNIST, XRMB, Flickr8k, Flickr30k and IAPR TC-12) on batches (750 samples per batch for MNIST and XRMB, 1024 for Flickr8k, Flickr30k and IAPR TC-12) using the RMSprop optimizer \cite{Tieleman2012} with a learning rate of $1e^{-3}$ and with a weight decay regularization of $1e^{-5}$.

After each training epoch, the loss of the current checkpoint of the model is computed. The learnable parameters that achieve the lowest loss on the validation set are chosen as the parameters of the model. The hyperparameters of each model are tuned by evaluating, on the validation set, the total canonical correlation and recall measurements for the total canonical correlation and retrieval experiments, respectively.

\begin{table}[H]
\centering
\begin{tabular}{@{}l@{~}|@{~}l@{~}l@{}} 
\toprule
Dataset   & $\mathbf{x}_1 / \mathbf{x}_2$  &Layers of $\mathbf{f}_{1}$/$\mathbf{f}_{2}$   \\ 
\hline
MNIST     & L/R   & 800,800,50/800,800,50\\
XRMB      & ACO/ART & 1800,1800,112/1200,1200,112 \\
Flickr\textsubscript{*}, IAPR  & IMG/TXT & 1024,512,128/1024,512,128   \\
\bottomrule
\end{tabular}
\caption{The neural network architectures of the mapping functions used for each dataset. L=left, R=right, ACO=acoustic, ART=articulatory, IMG=image, TXT=text}
\label{tbl:network_architectures}
\end{table}

Regarding our proposed DS-DCCA, we model the conventionally static layers of the mapping functions with neural networks, which are identical to the networks modelling the mapping functions in the network-based baselines, i.e. according to Table \ref{tbl:network_architectures}. In addition, we model the scaling networks with fully connected layers followed by a batch normalization and a ReLU activation function. The last layer, which outputs the scaling factors, is not followed by an activation function. The number of layers and neurons of the scaling networks are chosen out of $\{(128,c),(256,c),(256,128,c)\}$, where $c$ is equal to the total number of conventional static parameters to be scaled. The number of epochs defining the warm-up period, $T$, is set to $50$.

The regularization coefficients of the linear CCA were tuned in the logarithmic-scaled range of $[10^{-8}, 10^{2}]$. For FKCCA and NKCCA, following prior works \cite{lopez2014randomized,michaeli2016nonparametric}, a rank-6,000 approximation of the kernel matrices is used. The regularization coefficients of DCCA are chosen out of $\{1e^{-6},1e^{-4},1e^{-2}\}$. DCCAE was trained with the best performing regularization coefficient of DCCA and the reconstruction coefficient is chosen out of $\{1e^{-3}, 1e^{-2}, 1e^{-1}, 1\}$.
For NCCA, the dimensions of the input views are reduced by PCA, as suggested by \cite{michaeli2016nonparametric}. In particular, each view of XRMB was reduced to 20\% of its original dimension, as done in the original NCCA study. For training NCCA on MNIST and Flickr8k, the dimensions of each view were reduced to the minimal number of PCA components, which captures the variance of at least 80\% of the raw data.
Regarding Soft-CCA, the stochastic regularization coefficient ($\lambda_{SDL}$) and the running-average coefficient ($\alpha$) of each view are chosen out of $\{1, 2, 4, 5, 10, 15, 20\}$ and $\{0.85, 0.9, 0.95, 0.97\}$, respectively.
Regarding $\ell_{0}$-CCA, the gates' regularization term and the noise's standard deviation were chosen out of $\{0.01, 0.1, 0.5\}$ and $\{0.1, 0.5, 0.75, 1\}$, respectively. The stochastic gates of $\ell_{0}$-CCA were initialized by sampling a Gaussian distribution with a mean of 0.5 and a standard deviation of 0.01, as done in the official implementation of the stochastic gate. 

In the retrieval experiment, both Ranking-CCA and our DS-Ranking CCA were trained for 500 (200) epochs on Flickr8k (Flickr30 and IAPR TC-23). The running-average coefficient of the CCA layer of the Ranking-CCA and the margin used in the pairwise ranking loss ($m$ in Equation 7) were chosen out of $\{0.85, 0.9, 0.95, 0.97\}$ and $\{0.4, 0.5, 0.6, 0.7, 0.8\}$, respectively.

All neural network based models were implemented in Pytorch 1.4 \cite{NEURIPS2019_9015} and were trained on a single GPU (NVIDIA GEFORCE RTX 2080 Ti). FKCCA and NKCCA were trained using their official implementation \footnote{\href{https://github.com/lopezpaz/randomized_nonlinear_component_analysis}{https://github.com/lopezpaz/randomized\_nonlinear\_component\_analysis}} on a CPU (Intel(R) Xeon(R) Silver 4114). NCCA was trained using its official implementation \footnote{\href{https://tomer.net.technion.ac.il/files/2017/08/NCCAcode\_v3.zip}{https://tomer.net.technion.ac.il/files/2017/08/NCCAcode\v3.zip}} on a CPU (Intel(R) Core-i7).

\section{Multiple Runs With Different Initial Seeds}
\label{seq:multi_seeds}
\subsection{Total Canonical Correlation}
\label{subseq:tc_multi_seeds}
In order to examine the sensitivity of the models to the initial seed, we train each neural network based model with its best hyper-parameters (as obtained on the validation set, without considering the test data) for 5 different initial seeds. For each measurement, we report the average, standard deviation, maximum and minimum values. In addition, we provide the p-value computed in t-test vs. our proposed DS-DCCA.
The total canonical correlation statistics are provided in tables \ref{tbl:total_corr_multi_mnist}, \ref{tbl:total_corr_multi_xrmb} and \ref{tbl:total_corr_multi_flickr8k} for MNIST, XRMB and Flickr8k. 

Evidently, for all three datasets (MNIST, XRMB and Flickr8k), our proposed dynamically-scaled Deep CCA (DS-DCCA) learns representations of the two views, which are more canonically correlated than the baselines on average of the 5 different runs. In fact, the minimum result out of the 5 runs of our proposed dynamically-scaled models outperforms the maximum obtained results of the baselines in most of the experiments. The very low p-values emphasize that the leading results of our DS-DCCA are statically significant.

\begin{table}[H]
\centering
\begin{tabular}{l|rrrr} 
\toprule
\multirow{2}{*}{Model} & \multicolumn{4}{c}{MNIST}                                                                                                           \\ 
\cline{2-5}
                       & \multicolumn{1}{c|}{MEAN$\pm$STD} & \multicolumn{1}{c|}{P-VALUE} & \multicolumn{1}{c|}{MAX} & MIN   \\ 
\hline
Up. Bound            &50.00$\pm$0.000 &{-} &50.00 &50.00     \\  \hline
CCA &29.05$\pm$0.000 &7.53E-12 &29.05	&29.05 \\
FKCCA &41.68$\pm$0.018 &2.22E-10 &41.71	&41.66 \\
NKCCA &45.01$\pm$0.013 &5.06E-08 &45.04	&45.00 \\
DCCA            &46.70$\pm$0.012	&1.48E-06
 &46.72	&46.69	 \\ 
DCCAE           &46.71$\pm$0.012	&5.02E-06
 &46.73	&46.70 \\
NCCA &37.37$\pm$0.000 &8.29E-11 &37.37	&37.37 \\
Soft-CCA        &44.59$\pm$0.117	&1.37E-06
 &44.71	&44.42\\ 
Soft-HGR        &46.81$\pm$0.019	&5.79E-06
 &46.83	&46.78	\\ 
$\ell_{0}$-CCA  &47.09$\pm$0.019	&6.10E-06
 &47.12	&47.07	 \\ 
DS-DCCA         &\textbf{47.49}$\pm$0.044	&{-} &\textbf{47.56}	&\textbf{47.44}	  \\ 
\bottomrule
\end{tabular}
\caption{Total canonical correlation statistics for MNIST}
\label{tbl:total_corr_multi_mnist}
\end{table}

\begin{table}[H]
\centering
\begin{tabular}{l|rrrr} 
\toprule
\multirow{2}{*}{Model} & \multicolumn{4}{c}{XRMB}                                                                                                           \\ 
\cline{2-5}
                       & \multicolumn{1}{c|}{MEAN$\pm$STD} & \multicolumn{1}{c|}{P-VALUE} & \multicolumn{1}{c|}{MAX} & MIN  \\ 
\hline
Up. Bound            &112.00$\pm$0.000 &{-} &112.00 &112.00      \\  \hline
CCA &16.02$\pm$0.000 &1.50E-15 &16.02 &16.02 \\
FKCCA &95.17$\pm$0.073 &1.99E-10 &95.30	&95.11 \\
NKCCA &103.31$\pm$0.052 &4.93E-10 &103.35	&103.25 \\
DCCA            	&108.66$\pm$0.031 &5.78E-08
 &108.69	&108.62  \\ 
DCCAE           	&108.66$\pm$0.014 &1.51E-08

	&108.68	&108.64    \\
NCCA &107.31$\pm$0.004 &9.95E-10 &107.31	&107.30 \\
Soft-CCA       &83.65$\pm$3.183	&4.54E-05

 &87.32	&79.02 \\ 
Soft-HGR	&106.01$\pm$0.049 &4.37E-09

	&106.05	&105.92 \\ 
$\ell_{0}$-CCA  &108.66$\pm$0.010	&1.03E-08

 &108.68	&108.65  \\ 
DS-DCCA         	&\textbf{110.84}$\pm$0.027	&{-} &\textbf{110.88}	&\textbf{110.81}    \\ 
\bottomrule
\end{tabular}
\caption{Total canonical correlation statistics for XRMB}
\label{tbl:total_corr_multi_xrmb}
\end{table}

\begin{table}[H]
\centering
\begin{tabular}{l|rrrr} 
\toprule
\multirow{2}{*}{Model} & \multicolumn{4}{c}{Flickr8k}                                                         \\ 
\cline{2-5}
                       & \multicolumn{1}{c|}{MEAN$\pm$STD} & \multicolumn{1}{c|}{P-VALUE} & \multicolumn{1}{c|}{MAX} & MIN  \\ 
\hline
Up. Bound            &128.00$\pm$0.000 &{-} &128.00 &128.00      \\ 
\hline
CCA &41.60$\pm$0.000 &3.81E-09 &41.60	&41.60 \\
FKCCA &38.15$\pm$0.253 &9.92E-10 &38.57	&37.94 \\
NKCCA &67.35$\pm$0.067 &1.54E-07 &67.45	&67.29 \\
DCCA &67.72$\pm$0.441 &6.31E-07
 &68.24	&67.05 \\
DCCAE &67.65$\pm$0.365 &1.14E-06
	&68.11	&67.14 \\
NCCA &52.95$\pm$0.071 &9.07E-09 &53.01	&52.84 \\
Soft-CCA &60.52$\pm$0.468 &9.03E-08
	&61.24	&59.94 \\
Soft-HGR &55.18$\pm$2.886 &1.33E-05
	&57.97	&50.45 \\
$\ell_{0}$-CCA &72.60$\pm$0.394	&1.24E-06
 &73.24	&72.18 \\
DS-DCCA &\textbf{86.04}$\pm$0.499	&{-} &\textbf{86.76}	&\textbf{85.53} \\
\bottomrule
\end{tabular}
\caption{Total canonical correlation statistics for Flickr8k}
\label{tbl:total_corr_multi_flickr8k}
\end{table}

\subsection{Cross-Modality Retrieval}
\label{subseq:retrieval_multi_seeds}
We conduct a similar experiment for testing the sensitivity of the leading cross-modality retrieval models to the initialization. In particular, we train the best performing compared baseline, Ranking CCA (R.CCA), and the full-configuration ($\mathbf{z}$ \& $\mathbf{x}$) of our proposed DS-R.CCA for 15 different initial seeds. For each measurement, we report the average, standard deviation, maximum and minimum values. In addition, we provide the p-values computed in paired t-test vs. our proposed DS-R.CCA.
The p-values for Flickr8k, Flickr30k and IAPR TC-12 are provided in table \ref{tbl:retrieval_multi_pvalue}, the means and standard deviations are provided in table \ref{tbl:retrieval_multi_mean_std} and the maximum and minimum results are provided in table \ref{tbl:retrieval_multi_max_min}. 

Evidently, our proposed dynamically-scaled Ranking-CCA (DS-R.CCA) achieves better recall rates than the conventionally-scaled Ranking-CCA for the three datasets (Flickr8k, Flickr30k and IAPR TC-12) on average of the 15 runs. The low p-values emphasize that the results are statistically significant.

\begin{table*}
\centering
\begin{tabular}{l|l|l|ccc} 
\toprule
\multirow{2}{*}{Dataset}       & \multicolumn{1}{c|}{\multirow{2}{*}{Task}} & \multicolumn{1}{c|}{\multirow{2}{*}{Model}} & \multicolumn{3}{c}{P-VALUE}                                                            \\ 
\cline{4-6}
                               & \multicolumn{1}{c|}{}                      & \multicolumn{1}{c|}{}                       & \multicolumn{1}{c|}{$R_{1}$} & \multicolumn{1}{c|}{$R_{5}$} & \multicolumn{1}{c}{$R_{10}$}  \\ 
\hline
\multirow{4}{*}{Flickr8k}      & \multirow{2}{*}{I $\rightarrow$ T}                 
& R.CCA &7.64E-05 &1.12E-06 &2.51E-07 \\
& & DS-R.CCA &{-} &{-} &{-} \\ 
\cline{2-6}
                               & \multirow{2}{*}{T $\rightarrow$ I}             
& R.CCA &2.75E-06 &1.50E-08 &1.33E-08 \\
& & DS-R.CCA &{-} &{-} &{-} \\ 
\hline
\hline
\multirow{4}{*}{Flickr30k}     & \multirow{2}{*}{I $\rightarrow$ T}                  
& R.CCA &6.65E-04 &3.89E-07 &3.32E-04 \\
& & DS-R.CCA &{-} &{-} &{-} \\ 
\cline{2-6}
                               & \multirow{2}{*}{T $\rightarrow$ I}  
& R.CCA &1.49E-03 &1.89E-05 &1.40E-05 \\
& & DS-R.CCA &{-} &{-} &{-} \\ 
\hline
\hline
\multirow{4}{*}{IAPR TC-12} & \multirow{2}{*}{I $\rightarrow$ T} 
& R.CCA &1.77E-02 &2.89E-02 &3.09E-02 \\
& & DS-R.CCA &{-} &{-} &{-} \\ 
\cline{2-6}
                               & \multirow{2}{*}{T $\rightarrow$ I}
& R.CCA &2.04E-02 &3.81E-04 &4.36E-04 \\
& & DS-R.CCA &{-} &{-} &{-} \\ 
\bottomrule
\end{tabular}
\caption[Recall rates statistics: p-value of T-Test]{Recall rates statistics: p-value computed in t-test vs. our DS-R.CCA (I=image, T=text)}
\label{tbl:retrieval_multi_pvalue}
\end{table*}

\begin{table*}
\centering
\begin{tabular}{l|l|l|ccc} 
\toprule
\multirow{2}{*}{Dataset}       & \multicolumn{1}{c|}{\multirow{2}{*}{Task}} & \multicolumn{1}{c|}{\multirow{2}{*}{Model}} & \multicolumn{3}{c}{MEAN$\pm$STD}                                                            \\ 
\cline{4-6}
                               & \multicolumn{1}{c|}{}                      & \multicolumn{1}{c|}{}                       & \multicolumn{1}{c|}{$R_{1}$} & \multicolumn{1}{c|}{$R_{5}$} & \multicolumn{1}{c}{$R_{10}$}  \\ 
\hline
\multirow{4}{*}{Flickr8k}      & \multirow{2}{*}{I $\rightarrow$ T}                 
& R.CCA &33.2$\pm$1.065 &66.1$\pm$1.078 &78.4$\pm$0.995 \\
& & DS-R.CCA &\textbf{35.6}$\pm$0.989 &\textbf{69.3}$\pm$0.881 &\textbf{81.5}$\pm$0.887 \\ 
\cline{2-6}
                               & \multirow{2}{*}{T $\rightarrow$ I}             
& R.CCA &31.8$\pm$0.861 &64.3$\pm$0.876 &77.6$\pm$0.795 \\
& & DS-R.CCA &\textbf{33.7}$\pm$0.683 &\textbf{67.8}$\pm$0.795 &\textbf{80.7}$\pm$0.811 \\ 
\hline
\hline
\multirow{4}{*}{Flickr30k}     & \multirow{2}{*}{I $\rightarrow$ T}                  
& R.CCA &40.6$\pm$1.075 &73.0$\pm$0.933 &83.1$\pm$0.833 \\
& & DS-R.CCA &\textbf{42.4}$\pm$1.146 &\textbf{75.0}$\pm$0.595 &\textbf{84.4}$\pm$0.662 \\ 
\cline{2-6}
                               & \multirow{2}{*}{T $\rightarrow$ I}  
& R.CCA &40.6$\pm$0.804 &71.6$\pm$1.260 &82.5$\pm$0.735 \\
& & DS-R.CCA &\textbf{42.0}$\pm$0.898 &\textbf{73.9}$\pm$0.627 &\textbf{84.3}$\pm$0.664 \\ 
\hline
\hline
\multirow{4}{*}{IAPR TC-12} & \multirow{2}{*}{I $\rightarrow$ T} 
& R.CCA &48.1$\pm$0.947 &82.3$\pm$0.616 &90.3$\pm$0.502 \\
& & DS-R.CCA &\textbf{49.1}$\pm$0.958 &\textbf{82.7}$\pm$0.630 &\textbf{90.8}$\pm$0.572 \\ 
\cline{2-6}
                               & \multirow{2}{*}{T $\rightarrow$ I}
& R.CCA &49.2$\pm$0.791 &81.1$\pm$0.572 &89.7$\pm$0.685 \\
& & DS-R.CCA &\textbf{50.0}$\pm$0.794 &\textbf{81.9}$\pm$0.642 &\textbf{90.5}$\pm$0.504 \\ 
\bottomrule
\end{tabular}
\caption[Recall rates statistics: mean and standard deviation]{Recall rates statistics: mean and standard deviation (I=image, T=text)}
\label{tbl:retrieval_multi_mean_std}
\end{table*}

\begin{table*}
\centering
\begin{tabular}{l|l|l|ccc|ccc} 
\toprule
\multirow{2}{*}{Dataset}       & \multicolumn{1}{c|}{\multirow{2}{*}{Task}} & \multicolumn{1}{c|}{\multirow{2}{*}{Model}} & \multicolumn{3}{c|}{MAX}     & \multicolumn{3}{c}{MIN}                                                        \\ 
\cline{4-9}
                               & \multicolumn{1}{c|}{}                      & \multicolumn{1}{c|}{}                       & \multicolumn{1}{c|}{$R_{1}$} & \multicolumn{1}{c|}{$R_{5}$} & \multicolumn{1}{c|}{$R_{10}$} &
                               \multicolumn{1}{c|}{$R_{1}$} & \multicolumn{1}{c|}{$R_{5}$} & \multicolumn{1}{c}{$R_{10}$}  
                               
                               \\ 
\hline
\multirow{4}{*}{Flickr8k}      & \multirow{2}{*}{I $\rightarrow$ T}                 
& R.CCA &35.3 &67.9 &79.7 &31.8 &64.5 &75.7\\
& & DS-R.CCA &\textbf{37.0} &\textbf{70.9} &\textbf{83.2} &\textbf{33.1} &\textbf{67.6} &\textbf{80.0}\\ 
\cline{2-9}
                               & \multirow{2}{*}{T $\rightarrow$ I}             
& R.CCA &34.0 &65.7 &78.9 &30.2 &63.0 &75.8\\
& & DS-R.CCA &\textbf{34.9} &\textbf{69.3} &\textbf{82.3} &\textbf{32.3} &\textbf{66.9} &\textbf{79.6}\\ 
\hline
\hline
\multirow{4}{*}{Flickr30k}     & \multirow{2}{*}{I $\rightarrow$ T}                  
& R.CCA &42.5 &74.2 &84.5 &38.6 &70.9 &81.8\\
& & DS-R.CCA &\textbf{43.9} &\textbf{75.9} &\textbf{85.5} &\textbf{39.9} &\textbf{74.0} &\textbf{83.3}\\ 
\cline{2-9}
                               & \multirow{2}{*}{T $\rightarrow$ I}  
& R.CCA &42.5 &73.5 &84.2 &39.2 &70.0 &81.0\\
& & DS-R.CCA &\textbf{44.2} &\textbf{75.0} &\textbf{85.6} &\textbf{40.3} &\textbf{73.0} &\textbf{83.4}\\ 
\hline
\hline
\multirow{4}{*}{IAPR TC-12} & \multirow{2}{*}{I $\rightarrow$ T} 
& R.CCA &49.6 &83.4 &91.0 &46.7 &81.4 &89.1\\
& & DS-R.CCA &\textbf{50.8} &\textbf{84.1} &\textbf{91.7} &\textbf{47.5} &\textbf{81.9} &\textbf{90.0}\\ 
\cline{2-9}
                               & \multirow{2}{*}{T $\rightarrow$ I}
& R.CCA &50.6 &81.9 &90.6 &47.8 &80.1 &88.5\\
& & DS-R.CCA &\textbf{50.8} &\textbf{83.0} &\textbf{91.3} &\textbf{48.3} &\textbf{80.6} &\textbf{89.8}\\ 
\bottomrule
\end{tabular}
\caption[Recall rates statistics: maximum and minimum values]{Recall rates statistics: maximum and minimum values (I=image, T=text)}
\label{tbl:retrieval_multi_max_min}
\end{table*}

\end{document}